\documentclass[10pt,twocolumn,letterpaper]{article}

\usepackage{cvpr}              
\usepackage{multirow}
\usepackage{pifont}
\usepackage[accsupp]{axessibility}

\usepackage[dvipsnames]{xcolor}

\definecolor{cvprblue}{rgb}{0.21,0.49,0.74}
\usepackage[pagebackref,breaklinks,colorlinks,citecolor=cvprblue]{hyperref}

\definecolor{purple}{rgb}{0.56,0.27,0.68}
\definecolor{red}{rgb}{0.95,0.4,0.4}
\definecolor{purered}{rgb}{1,0,0}
\definecolor{blue}{rgb}{0.4,0.4,0.95}
\definecolor{darkblue}{rgb}{0,0,0.8}
\definecolor{grey}{rgb}{0.6,0.6,0.6}
\definecolor{col1}{RGB}{232, 161, 148}
\definecolor{col2}{RGB}{148, 187, 232}
\definecolor{col3}{RGB}{206, 239, 255}
\definecolor{lightgrey}{rgb}{0.85,0.85,0.85}
\definecolor{lightlightgrey}{rgb}{0.9,0.9,0.9}
\definecolor{verylightBG}{rgb}{0.9,0.99,0.99}
\definecolor{darkgreen}{rgb}{0.3, 0.75, 0.3}
\definecolor{pink}{RGB}{242,78,174}

\def\paperID{1618} 
\def\confName{CVPR}
\def\confYear{2024}

\title{GCT: Graph Co-Training for Semi-Supervised Few-Shot Learning}

\author{Rui Xu\textsuperscript{\rm 1}, 
Lei Xing\textsuperscript{\rm 1}, 
Shuai Shao\textsuperscript{\rm 1}, 
Lifei Zhao\textsuperscript{\rm 1}, 
Baodi Liu\textsuperscript{\rm 1}, 
Weifeng Liu\textsuperscript{\rm 1},
Yicong Zhou\textsuperscript{\rm 2}\\
\textsuperscript{\rm 1}China University of Petroleum (East China),
\textsuperscript{\rm 2}University of Macau
\\
{\tt\small thu.liubaodi@gmail.com, liuwf@upc.edu.cn}
}

\begin{document}
\maketitle

\begin{abstract}
Few-shot learning (FSL), purposing to resolve the problem of data-scarce, has attracted considerable attention in recent years. A popular FSL framework contains two phases: (i) the pre-train phase employs the base data to train a CNN-based feature extractor. (ii) the meta-test phase applies the frozen feature extractor to novel data (novel data has different categories from base data) and designs a classifier for recognition. To correct few-shot data distribution, researchers propose Semi-Supervised Few-Shot Learning (SSFSL) by introducing unlabeled data. Although SSFSL has been proved to achieve outstanding performances in the FSL community, there still exists a fundamental problem: the pre-trained feature extractor can not adapt to the novel data flawlessly due to the cross-category setting. Usually, large amounts of noises are introduced to the novel feature. We dub it as Feature-Extractor-Maladaptive (FEM) problem. To tackle FEM, we make two efforts in this paper. First, we propose a novel label prediction method, Isolated Graph Learning (IGL). IGL introduces the Laplacian operator to encode the raw data to graph space, which helps reduce the dependence on features when classifying, and then project graph representation to label space for prediction. The key point is that: IGL can weaken the negative influence of noise from the feature representation perspective, and is also flexible to independently complete training and testing procedures, which is suitable for SSFSL. Second, we propose Graph Co-Training (GCT) to tackle this challenge from a multi-modal fusion perspective by extending the proposed IGL to the co-training framework. GCT is a semi-supervised method that exploits the unlabeled samples with two modal features to crossly strengthen the IGL classifier. We estimate our method on five benchmark few-shot learning datasets and achieve outstanding performances compared with other state-of-the-art methods. It demonstrates the effectiveness of our GCT.

\end{abstract}

\section{Introduction}
\label{sec: introduction}

\begin{figure}[t]
	\begin{center}
		\includegraphics[width=1.0\linewidth]{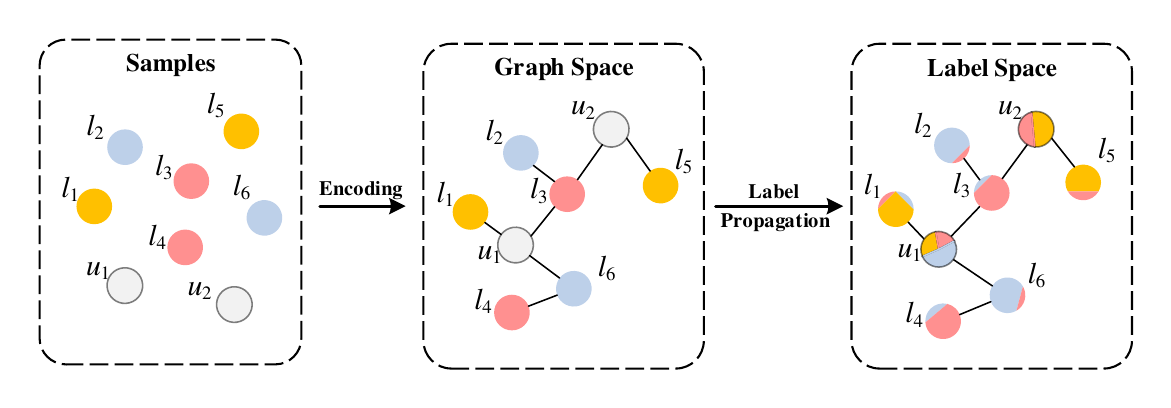}
	\end{center}
	\caption{Isolated Graph Learning (IGL) classifier. $l_{\cdot}$ and $u_{\cdot}$ denote the labeled and unlabeled samples, respectively. IGL first encodes the samples to graph representation and then propagate the label information through graph structure for prediction.}
	\label{figure: IGL}
\end{figure}

\begin{figure*}[t]
	\begin{center}
		\includegraphics[width=1.0\linewidth]{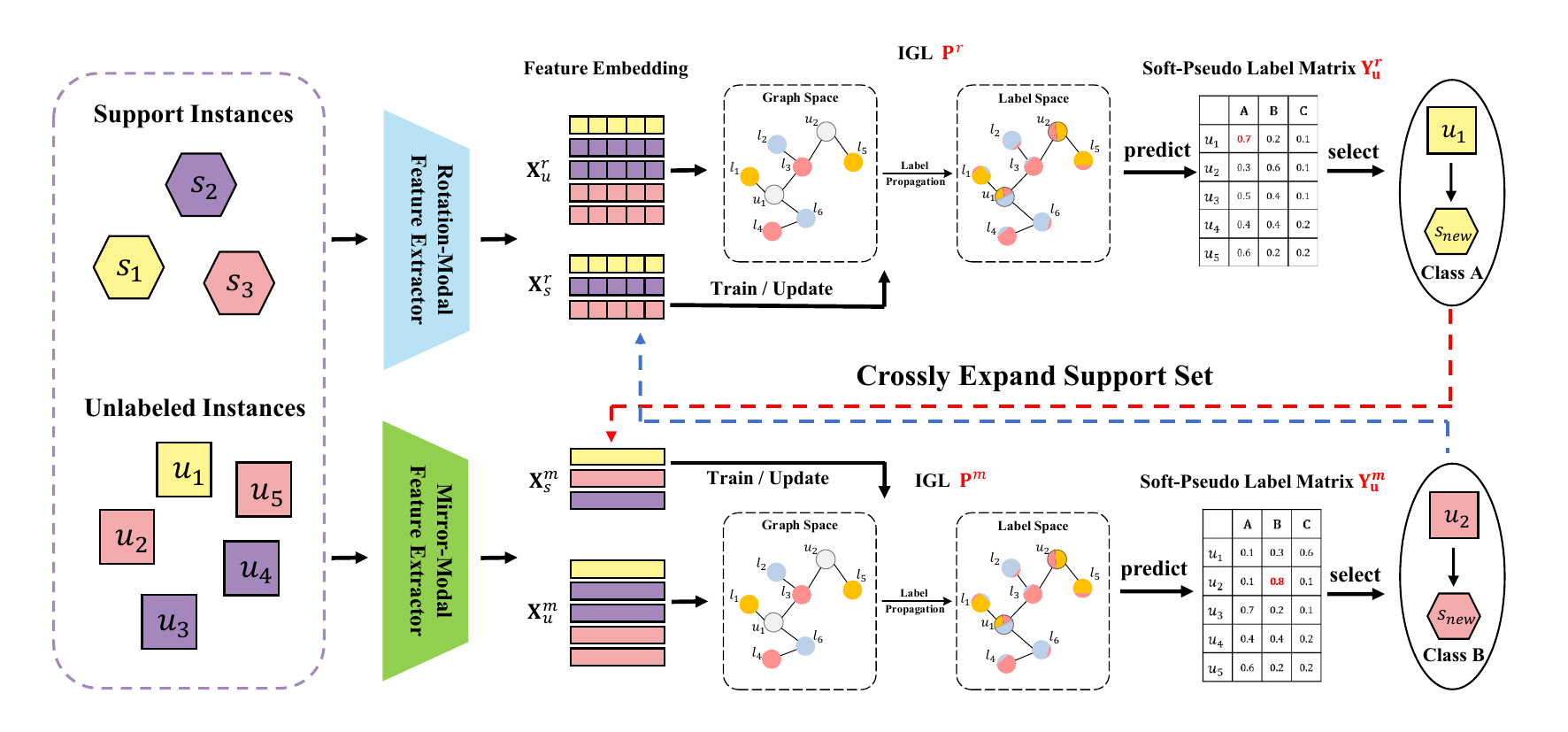}
	\end{center}
	\caption{The flowchart of Graph Co-Training (GCT) in inductive semi-supervised case. We have two kinds of feature extractors, \textit{i.e.}, rotation-modal feature extractor and mirror-modal feature extractor. $\mathbf{X}_s^r$ and $\mathbf{X}_s^m$ indicate the features of support data in rotation-modality and mirror-modality. $\mathbf{X}_u^r$ and $\mathbf{X}_u^m$ represent the features of unlabeled data in corresponding modalities.
	In each modality, we first employ the support samples to train the basic classifiers $\mathbf{P}^{r}$ and $\mathbf{P}^{m}$ (\textit{i.e.}, IGL). Then, we use IGL classifier to predict unlabeled samples and obtain the corresponded soft-pseudo label matrices $\mathbf{Y}_u^{r}$ and $\mathbf{Y}_u^{m}$. 
	Next, we select the most confident sample and give it one-hot-pseudo label. Note that different modalities may predict different results.
	Finally, we crossly expand the predicted-pseudo-sample to the support set and update the IGL classifier.}
	\label{figure: Flowcharts}
\end{figure*}

In recent years, the performance of computer vision tasks based on deep learning has reached or even surpassed the human beings' level, such as image classification \cite{shao2021mdfm,wang2016cost,shao2020label}, person re-identification \cite{wang2020dense,zheng2018pedestrian,fan2020contextual}, and point cloud recognition \cite{wang2022rotation,qian2021assanet,zhao2022real}.
The adequate labeled data plays a crucial role for the success. However, it is a challenge for data collection and maintenance in real-world situations.
To this end, few-shot learning (FSL), as a pioneer work to address the lack of labeled samples for each category, has aroused widespread concerns.

In a standard FSL, the employed data includes two parts, \textit{i.e.}, base set and novel set.
There are many labeled samples in the base set, but very few in the novel set (typically, for the general FSL setting, each category only has $1$ or $5$ labeled samples).
Notably, the categories contained in the base set are entirely different from those in the novel set.
Generally, researchers split the FSL model into two phases:
(i) pre-train. Training a feature extractor through the base set. 
(ii) meta-test. First, employing the feature extractor to extract the features of novel data, and then designing a classifier to recognize the novel data's category.
Besides, to overcome overfitting caused by the \textit{few-shot} setting, researchers prefer to decouple the complete model, that is, freezing the parameters of the feature extractor after pre-training and directly extracting the cross-category novel features in the meta-test phase.

\begin{table*}[t]
        \caption{The definition of abbreviations and notations.}
        \begin{center}
        \resizebox{\textwidth}{!}{
            \begin{tabular}{ll}
                \toprule
                \textbf{Abbreviation and Notation}                  & \textbf{Definition}      \\
                \midrule
                IGL          & Isolated Graph Learning \\
                GCT          & Graph Co-Training \\
                FEM          & feature-extractor-maladaptive \\
                Std-Mod      & standard modality \\
                Meta-Mod     & meta modality \\
                SS-R-Mod     & self-supervised rotation modality \\
                SS-M-Mod     & self-supervised mirror modality \\
                FSL          & few-shot learning \\
                ISFSL         & inductive supervised few-shot learning \\
                TSFSL         & transductive supervised few-shot learning \\
                ISSFSL        & inductive semi-supervised few-shot learning \\
                TSSFSL        & transductive semi-supervised few-shot learning \\
                \midrule
                $\mathcal{D}_{base}$, $\mathcal{D}_{novel}$  & base data, novel data \\
                $\mathcal{S}$, $\mathcal{Q}$, $\mathcal{U}$ & support set, query set, unlabeled set \\
                $\omega^r(\cdot)$, $\omega^m(\cdot)$ & rotation-modal feature extractor, mirror-modal feature extractor \\
                $\mathbf{A}$ & adjacency matrix \\
                $\mathbf{D}$ & vertex degree matrix \\
                
                $\mathbf{X}$ & feature embedding of training data\\
                $\mathbf{x}_{ts}$ & feature embedding of testing data\\
                $\mathbf{X}_{s}^r$, $\mathbf{X}_{s}^m$ & support feature embedding on rotation-modal and mirror-modal\\
                $\mathbf{X}_{u}^r$, $\mathbf{X}_{u}^m$ & unlabeled feature embedding on rotation-modal and mirror-modal\\                
                $\mathbf{X}_{q}^r$, $\mathbf{X}_{q}^m$ & query feature embedding on rotation-modal and mirror-modal\\
                $\mathbf{x}_{select}^r$, $\mathbf{x}_{select}^m$ & selected the most confidence samples' feature embedding on rotation-modal and mirror-modal\\
                
                $\mathbf{Y}$ & initial label matrix of training data\\
                $\mathbf{Y}_s^{r}$, $\mathbf{Y}_s^{m}$ & one-hot label matrices of support data on rotation-modal and mirror-modal\\                
                $\mathbf{Y}_u^{r}$, $\mathbf{Y}_u^{m}$ & predicted soft-pseudo label matrices of unlabeled data on rotation-modal and mirror-modal\\
                $\mathbf{Y}_q^{r}$, $\mathbf{Y}_q^{m}$ & predicted soft label matrices of query data on rotation-modal and mirror-modal\\                
                $\mathbf{y}_{select}^r$, $\mathbf{y}_{select}^m$ & selected the most confidence samples' one-hot-pseudo label vectors on rotation-modal and mirror-modal\\   
                
                $\mathbf{P}$ & classifier \\
                $\mathbf{P}^r$, $\mathbf{P}^m$ & rotation-modal classifier, mirror-modal classifier\\
                $\mathbf{P}_{opt}^r$, $\mathbf{P}_{opt}^m$ & optimal rotation-modal classifier, optimal mirror-modal classifier\\
              \bottomrule
            \end{tabular} 
        }
        \end{center}
        \label{table: definition}        
\end{table*}

During the the stage of designing the classifier, the FSL-based methods can be categorized into two sorts according to the type of data employed: (i) supervised FSL, and (ii) semi-supervised FSL.
Specifically, the novel data includes three components: support data (\textit{i.e.}, labeled training data), unlabeled data (\textit{i.e.}, unlabeled training data), and query data (\textit{i.e.}, to-be-classified testing data).
The difference between the two settings is whether to use the unlabeled data when building the classifier. For more details, please see Section \ref{sec:Problem Set-Up}.

Compared with the supervised FSL,
semi-supervised approaches \cite{li2019learning,wang2020instance,huang2021pseudo,lazarou2021iterative} can effectively correct few-shot data distributions to make the learned classifiers have higher quality.
However, to achieve this goal, it must be on the premise that sample features are less noisy.
But unfortunately, a fundamental problem in FSL, Feature-Extractor-Maladaptive (FEM), is easy to break up the assumption.
Specifically, researchers obtain a feature extractor in the pre-train process and apply it directly to the meta-test process, which is challenging to ensure that the frozen feature extractor is capable of adapting to the novel categories.
To solve this challenge, we make two efforts in this paper.

First, inspired by \cite{belkin2002laplacian}, we know that transforming the raw data to graph representation is helpful in reducing the dependence on features in classification tasks. 
To this end, we propose a novel label prediction method dubbed as Isolated Graph Learning (IGL), try to weaken the negative impact of noise from the feature representation perspective.
The framework of IGL is shown in Figure \ref{figure: IGL}.
We first introduce the graph Laplacian operator to encode the sample's feature embedding to graph space and then project the graph representation to label space for prediction by regularization.
Compared with the traditional graph learning method \cite{zhou2003learning}, needing both labeled and unlabelled data to propagate label information, our IGL is more flexible to independently complete training and testing procedures.
Furthermore, compared with graph neural network (GNN) based methods, there are no abundant parameters in our IGL that need to be updated with the propagation of deep neural networks. In other words, it is easy to be implemented.
These attributes are very friendly for few-shot classification in the semi-supervised setting.

Second, we propose Graph Co-Training (GCT) to weaken the negative effect of noise from a multi-modal fusion perspective. 
Suppose that we have features from different feature extractors, we can integrate different predictions (obtained from different features) through collaborative training (co-training) to complete the final classification.
To be more specific, we first try to get two-modal features from two designed feature extractors.
In this phase, we have the flexibility to adapt the classical models in few-shot communities.
This paper uses two kinds of self-supervision ways from \cite{mangla2020charting}, \cite{shao2021mhfc}.
Then, we exploit the support data to train two basic classifiers (\textit{i.e.}, IGL) with different modal features. Next, we separately predict the unlabeled data with the two modalities of classifiers.
At last, we apply the unlabeled data with the most confident predictions to crossly update the classifiers.
The designed co-training way is mainly to strengthen our classifier's robustness to reduce the interference caused by FEM.
We illustrate the framework of our GCT in Figure \ref{figure: Flowcharts}. For convenience, we list some crucial abbreviations and notations in Table \ref{table: definition}.

In summary, the main contributions focus on:
\begin{itemize}
\item
Aiming at the Feature-Extractor-Maladaptive (FEM) problem in semi-supervised few-shot learning, we propose a novel graph learning based classifier dubbed as Isolated Graph Learning (IGL), which completes training and testing procedures independently.
\item
We combine our IGL with the co-training framework to design a Graph Co-training (GCT) algorithm. It extends IGL to semi-supervised few-shot learning through fusing multi-modal information.
\item
The comparison results with SOTAs on five benchmark FSL datasets have evaluated the efficiency of our GCT.

\end{itemize}

\section{Related Work}
\label{Section: Related work}
\subsection{Semi-Supervised Few-Shot Learning}
Recently, semi-supervised few-shot learning (SSFSL) has attracted lots of attention. Researchers assume that abundant unlabeled data is available to be used for constructing the classifier. 
They introduce various traditional semi-supervised learning methods to the few-shot learning (FSL) task. Here, we list several classical semi-supervised approaches and corresponding FSL works.
(i) Consistency regularization methods aim to improve the robustness of classifiers when the images are noisy. MetaMix \cite{chen2020metamix}, BR-ProtoNet \cite{huang2020behavior} \emph{et.al.} promote the classifier from this way.
(ii) Self-training methods first train a classifier with labeled data, then exploit it to generate pseudo labels for unlabeled data, and at last update the classifier with pseudo-labeled data. Recent self-training based FSL methods, including LST \cite{li2019learning} ICI \cite{wang2020instance}, PLCM \cite{huang2021pseudo}, iLPC \cite{lazarou2021iterative} have achieved outstanding performances.
(iii) Hybrid based FSL methods, such as MixMatch \cite{berthelot2019mixmatch}, and FixMatch \cite{sohn2020fixmatch}, try to construct a unified framework of semi-supervised learning by combining several current dominant approaches such as self-training and consistency regularization.

\subsection{Multi-Modal Few-Shot Learning}
As there are two sides to every coin, it is boundedness to define objects from a single point of view. Multi-view learning as an effective strategy has attracted extensive attention in the past decade.
In few-shot learning, some similar methods have been proposed, such as:
DenseCls\cite{lifchitz2019dense}, its feature map is divided into various blocks, and the corresponding labels are predicted;
DivCoop \cite{dvornik2020selecting} trains the feature extractors on various datasets and integrates them into a multi-domain representation; 
URT \cite{liu2021universal} is an improved method compared with DivCoop \cite{dvornik2020selecting}, which proposes a transformer layer to help the network employ various datasets;
DWC \cite{dvornik2019diversity} introduces a cooperate strategy on a designed ensemble model to integrate multiple information.
Although the above-mentioned approaches are based on multi-modal learning, they are limited by the fused feature extractors and classifiers. 
To be more specific, their methods are based on the unified framework, which means that the feature extractors are usually guaranteed to match classifiers in a fixed way. Without this combination, the model performance will be significantly reduced, which greatly limits the scalability of the methods. While our GCT is freed with the feature extractor, thereby is more flexible to be applied in real scenarios. 

\subsection{Graph Learning}
\label{subsec: graph learning}
Graph Learning is an efficient way to model the data correlation of samples, composed of vertex set (\textit{i.e.}, samples) and edge set. Each edge connects two vertices, which is capable of modeling pair-wise relations of samples. Researchers usually employ the adjacency matrix to represent a graph structure.
\cite{zhou2003learning} first proposed Graph Learning. 
In recent years, graph-based neural networks (GNN) have received extensive attention and have developed rapidly. 
Due to its good performance, GNN-based technologies have been applied in many fields, including few-shot learning  \cite{garcia2017few,kim2019edge,yang2020dpgn,tang2021mutual}, face clustering \cite{yang2020learning,wang2019linkage}, etc.
The former achieves satisfactory performance by cooperating with meta-learning strategy, and the latter designs multiple kinds of GNNs to cooperate with each other to complete goals.

This paper focuses on traditional graph learning.
In graph learning, researchers classify the unlabelled data by propagating the label information through graph structure, which is constructed by employing all the data (including labeled and unlabeled data). 
There are two challenges in the node/graph classification task by using graph learning: (i) In real applications, it is hard to know in advance what the to-be-tested sample looks like. It limits the scenario for this approach.
(ii) Leading to a high computational cost. The predicting process is entirely online. All the data must be considered during the learning stage, which results in consuming massive computing resources. 
Furthermore, researchers have to re-construct the graph structure to propagate the label information when coming new testing data. 
This paper proposes a novel method dubbed as Isolated Graph Learning (IGL) to solve this challenge. IGL is a classifier which can be directly applied in the decoupled FSL task.

\section{Problem Setup}
\label{sec:Problem Set-Up}
In few-shot learning, there exist two processes (\textit{i.e.}, pre-train process and meta-test process) with different categories of samples. 
Define the base data in pre-train phase as $\mathcal{D}_{base} = \{(x_{(i)},y_{(i)}) | y_{(i)} \in \mathcal{C}_{base} \}_{i=1}^{N_{base}}$, and the novel data in meta-test phase as $\mathcal{D}_{novel} = \{(x_{(j)},y_{(j)}) | y_j \in \mathcal{C}_{novel} \}_{j=1}^{N_{novel}}$, where $x$ and $y$ denote the sample and corresponded label. $\mathcal{C}_{base}$ and $\mathcal{C}_{novel}$ indicate the categories of base data and novel data, $\mathcal{C}_{base} {\kern 2pt} \cap {\kern 2pt} \mathcal{C}_{novel} = \emptyset$. $N_{base}$ and $N_{novel}$ indicate the number of base data and novel data.
To overcome overfitting, we follow the decoupled classification setups as \cite{wang2020instance}.
There are three main stages in few-shot learning.
First, we use $\mathcal{D}_{base}$ to train a CNN-based feature extractor $\omega \left(  \cdot  \right)$ in the pre-train phase.
Second, we freeze the model's parameters and extract the feature embedding of $\mathcal{D}_{novel}$. Third, we design a classifier for the novel categories' classification. 

Specifically, we first define the meta-test dataset as $\mathcal{D}_{novel} = \{\mathcal{S}, \mathcal{Q}, \mathcal{U}\}$, where $\mathcal{S}$, $\mathcal{Q}$, and $\mathcal{U}$ indicate support set, query set, and unlabeled set,  $\mathcal{S} {\kern 2pt} \cap {\kern 2pt} \mathcal{Q} {\kern 2pt} = \emptyset$, $\mathcal{S} {\kern 2pt}  \cap {\kern 2pt} \mathcal{U} = \emptyset$, $\mathcal{Q} {\kern 2pt} \cap {\kern 2pt} \mathcal{U} = \emptyset$. 
Next, we divide the support, query, and unlabeled sets into different episodes. Each episode has $K$-$way$-$O$-$shot$ samples, where $K$-$way$ indicates $K$ classes, and $O$-$shot$ denotes $O$ samples per class.  
Finally, we employ the to-be-learned classifier to obtain the average classification accuracies of all the meta-test episodes on the query set $\mathcal{Q}$.

Besides, according to the differences of data used to construct the classifier, we split the few-shot learning into four kinds of settings:  (i) inductive supervised few-shot learning (ISFSL), using the support features and labels to train the classifier; (ii) transductive supervised few-shot learning (TSFSL), using the support features, support labels, and query features to train the classifier; (iii) inductive semi-supervised few-shot learning (ISSFSL), using the support features, support labels, and unlabeled features to train the classifier; (iv) transductive semi-supervised few-shot learning (TSSFSL), using the support features, support labels, unlabeled features, and query features to train the classifier.


\section{Methodology}
In this section, we describe our approach in detail.
First, we encode the graph structure of samples' relations to the adjacency matrix.
Second, we propose a novel graph learning method dubbed as Isolated Graph Learning (IGL) to tackle the FEM to some extent. This strategy introduces the Laplacian operator to transform the samples in feature space to graph representation and then project them to label space for prediction by regularization.
Some details are shown in Figure \ref{figure: IGL}.
Third, we propose Graph Co-Training (GCT) by expanding the IGL to a co-training framework.
GCT is capable of further addressing the FEM problem from a multi-modal fusion perspective.
The flowchart is illustrated in Figure \ref{figure: Flowcharts}.
At last, we introduce our designed feature extractors in detail.

\label{sec:Methodology}
\subsection{Graph Structure Encoder}
A graph can be formulated as $\mathcal{G} = (\mathcal{V}, \mathcal{E})$, where $\mathcal{V}$ and $\mathcal{E}$ represent vertex set and edge set, respectively. 
In the paper, the vertex set is composed of image samples.
We encode the relations between edges and vertices through adjacency matrix $\mathbf{A}\in{\mathbb{R}}^{|\mathcal{V}| \times |\mathcal{V}|}$.
Given the feature embedding of labeled vertices as $\mathbf{X} = [\mathbf{x}_{(1)}, \mathbf{x}_{(2)}, \cdots, \mathbf{x}_{(N)}] \in \mathbb{R}^{dim \times N}$, where $\mathbf{x}_{(i)},(i=1,2,\cdots,N)$ indicates the embedding of $v_i$, and $v_i$ denotes the $i_{th}$ vertex in $\mathcal{V}$.
$dim$ and $N$ denote the dimension and number of labeled samples.
The elements in $\mathbf{A}$ can be defined as:
\begin{equation}
\begin{split}
        \mathbf{A}_{(i,j)}=
        \left\{\begin{array}{cl}
            {\exp \left(-dis\left( \mathbf{x}_{(i)}, \mathbf{x}_{(j)}\right)^{2}\right)}  & {\text { if } (v_{(i)},v_{(j)}) \in e} \\
            {0} & {\text{ o.w. }}
        \end{array}\right.
\end{split}
\label{eqa:elements_in_A}
\end{equation}
where 
$(\cdot)_{(i,j)}$ is the (i,j)-element in $(\cdot)$.
$e$ indicates an edge in $\mathcal{E}$.  
$dis\left( \mathbf{x}_{(i)}, \mathbf{x}_{(j)}\right)$ represents the operator to calculate the distance of feature embeddings between $v_{(i)}$ and $v_{(j)}$, in our method, we select the $k$ Nearest Neighbor (KNN) method. 
Following, we define the vertex degree matrix as $\mathbf{D} \in \mathbb{R}^{N \times N}$, which denotes a diagonal matrix with its $(i,i)$-element equal to the sum of the $i$-${th}$ row of $\mathbf{A}$.

\subsection{Isolated Graph Learning}
In this section, we propose a novel label prediction method dubbed as Isolated Graph Learning (IGL). IGL is a strategy to solve the FEM problem by transforming the samples in feature space to graph space.
Unlike traditional graph learning, requiring both labeled and unlabelled data to construct the graph, the proposed IGL is more flexible to independently complete training and testing procedures by learning a regularized projection $\mathbf{P} \in \mathbb{R}^{dim \times C}$ to classify different categories. Here, $C$ indicates the total number of classes.
We calculate the cost function as:
\begin{equation}
\begin{split}
        \mathcal{F}(\mathbf{P}) = f_1(\mathbf{P}) + \lambda f_2(\mathbf{P})  + \mu f_3(\mathbf{P})
\end{split}
\label{eqa:cost_function}
\end{equation}
where $\lambda$ and $\mu$ represent the parameters to balance the function.
$f_1(\mathbf{P})$ denotes the graph Laplacian regularizer, which can be formulated as:
\begin{equation}
\begin{split}
        f_1(\mathbf{P}) 
        &= \frac{1}{2} \left(
        \sum_{i,j=1}^{N} \mathbf{A}_{(i,j)} 
        \left( 
        \frac{ \left( {\mathbf{X}}^T\mathbf{P}  \right)_{(i \cdot)} }{\sqrt{\mathbf{D}_{(i,i)}}} 
        - \frac{ \left( {\mathbf{X}}^T\mathbf{P} \right)_{(j \cdot)} }{\sqrt{\mathbf{D}_{(j,j)}}}
        \right)^2
        \right)\\
        &= \text{tr} \left( \mathbf{P}^T{\mathbf{X}} \mathbf{\Delta} {\mathbf{X}}^T \mathbf{P} \right)
\end{split}
\label{eqa:graph_laplacian}
\end{equation}
where $(\cdot)_{(i,\cdot)}$ is the $i$-${th}$ row of $(\cdot)$. 
$\mathbf{\Delta}= \mathbf{D}^{- \frac{1}{2}} \mathbf{A} \mathbf{D}^{- \frac{1}{2}}$ denotes the normalized graph Laplacian operator.
$f_2(\mathbf{P})$ indicates the empirical loss term, which can be formulated as:
\begin{equation}
\begin{split}
        f_2(\mathbf{P}) = \left \| {\mathbf{X}}^T \mathbf{P} - \mathbf{Y} \right\|_F^2 
\end{split}
\label{eqa:empirical_loss_term}
\end{equation}
where $\mathbf{Y} \in \mathbb{R}^{N \times C}$ indicates the initial label embedding matrix. For labeled samples, if the $i$-${th}$ sample belongs to the $j$-${th}$ class, $\mathbf{Y}_{(i,j)}$ is $1$ , and otherwise, it is $0$. $f_3(\mathbf{P})$ is the constraint term. In this paper, we introduce $\ell_{2,1}$-$norm$ to select an essential feature and avoid overfitting for $\mathbf{P}$, which can be defined as:
\begin{equation}
\begin{split}
        f_3(\mathbf{P}) = \left \| \mathbf{P} \right\|_{\ell_{2,1}} 
\end{split}
\label{eqa:l_21 constant}
\end{equation}
where $\left\| \cdot \right\|_{\ell_{2,1}}$ represents $\ell_{2,1}$-norm of $\left( \cdot \right)$.
The objective function on IGL can be formulated as:
\begin{equation}
\begin{split}
        &\mathop {\arg \min}\limits_{\mathbf{P}}     
        \mathcal{F}(\mathbf{P}) \\
        &= 
        \text{tr} \left( \mathbf{P}^T{\mathbf{X}} \mathbf{\Delta} {\mathbf{X}}^T \mathbf{P} \right)
        + \lambda \left \| {\mathbf{X}}^T \mathbf{P} - \mathbf{Y} \right\|_F^2
        + \mu \left \| \mathbf{P} \right\|_{\ell_{2,1}}
\end{split}
\label{eqa: Objective_function}
\end{equation}
To optimize this problem, we first relax Equation \ref{eqa: Objective_function} as:
\begin{equation}
\begin{split}
        &\mathop {\arg \min}\limits_{\mathbf{P},\mathbf{B}}     
        \mathcal{F}(\mathbf{P},\mathbf{B}) \\
        &= \text{tr} \left( \mathbf{P}^T{\mathbf{X}} \mathbf{\Delta} {\mathbf{X}}^T \mathbf{P} \right)
        + \lambda \left \| {\mathbf{X}}^T \mathbf{P} - \mathbf{Y} \right\|_F^2
        + \mu \text{tr}\left ( \mathbf{P}^T \mathbf{B} \mathbf{P} \right)
\end{split}
\label{eqa: relax_Objective_function}
\end{equation}
where $\mathbf{B}$ is a diagonal matrix. Then we alternately update $\mathbf{P}$ and $\mathbf{B}$ until Equation \ref{eqa: relax_Objective_function} convergence, follow \cite{zhang2018inductive}, we directly solve the problems as:
\begin{equation}
\begin{split}
        \mathbf{B}_{(i,i)} = \frac{1}{2 \left\| \mathbf{P}_{(i,\cdot)} \right\|^2_2 + 10^{-8}} , {\kern 4pt} i = 1,\dots,dim
\end{split}
\label{eqa: update_B}
\end{equation}
\begin{equation}
\begin{split}
        \mathbf{P}
        =\mathcal{H}(\mathbf{X})
        = \lambda \left( 
        {\mathbf{X}} \mathbf{\Delta} {\mathbf{X}}^T 
        + \lambda  {\mathbf{X}} {\mathbf{X}}^T
        + \mu \mathbf{B}
        \right)^{-1}
        {\mathbf{X}} \mathbf{Y}
\end{split}
\label{eqa: update_P}
\end{equation}
Following, given a testing sample embedding $\mathbf{x}_{ts} \in \mathbb{R}^{dim \times 1}$, we predict the $\mathbf{x}_{ts}$'s category by:
\begin{equation}
\begin{split}
        \mathcal{Z}(\mathbf{x}_{ts}) = id_{max}
        \left\{
        {\mathbf{x}_{ts}}^T \mathbf{P}  
        \right\}
\end{split}
\label{eqa: single_modal_predict}
\end{equation}
where $id_{max}$ represents an operator to obtain the index of the max value in the vector.

\subsection{Graph Co-Training for Few-Shot Learning}
\label{subsec: GCT}

As mentioned in Section \ref{sec:Problem Set-Up}, there exist four kinds of setting in FSL. 
To make our IGL perform well in all kinds of FSL, we introduce the co-training strategy to further cooperate with IGL, and named the new approach as Graph Co-Training (GCT).
On the one hand, it can solve the FEM problem from a multi-modal fusion perspective. For the details of multi-modal information, please refer to Section \ref{subsec: FE}.
On the other hand, GCT is capable of strengthening the robustness of the to-be-learned classifier by employing the unlabeled data $\mathcal{U}$.
We'll detail how GCT works in the four settings. 
Notably, both the construction of $\mathbf{P}$ and the collaborative training are designed for each episode.

First see the inductive semi-supervised few-shot learning (ISSFSL).
We construct the co-training framework with two modal features, \textit{i.e.}, rotation-modality and mirror-modality. 
The rotation-model feature extractor, $\omega^r(\cdot)$ follows \cite{mangla2020charting}, the corresponding embeddings of novel data can be defined as $\mathbf{X}_{novel}^r = \omega^r(\mathcal{D}_{novel}) = [{\mathbf{X}_s^r}, {\mathbf{X}_u^r}, {\mathbf{X}_q^r}]$, where ${\mathbf{X}_s^r}=\omega^r(\mathcal{S})$, ${\mathbf{X}_u^r}=\omega^r(\mathcal{U})$, and ${\mathbf{X}_q^r}=\omega^r(\mathcal{Q})$ indicate the features of support, unlabeled, and query data on the rotation-model.
The mirror-modal feature extractor, $\omega^m(\cdot)$ follows \cite{shao2021mhfc}, the features in this modal are denoted as $\mathbf{X}_{novel}^m = \omega^m(\mathcal{D}_{novel}) = [{\mathbf{X}_s^m}, {\mathbf{X}_u^m}, {\mathbf{X}_q^m}]$. The complete GCT is demonstrated in Figure \ref{figure: Flowcharts}, which consists of four steps:

  
    
    

(i) From Equation \ref{eqa: update_P}, we construct two different classifiers $\mathbf{P}^r$ and $\mathbf{P}^m$ by employing two modal support features $\mathbf{X}^r_{s}$ and $\mathbf{X}^m_{s}$, respectively.
\begin{equation}
\begin{split}
        \left\{\begin{array}{ll}
            \mathbf{P}^r 
            = \mathcal{H}(\mathbf{X}_s^r)\\
            \mathbf{P}^m 
            = \mathcal{H}(\mathbf{X}_s^m)
        \end{array}\right.
\end{split}
\label{eqa: two_model_base_learner}
\end{equation}

(ii) Predict the unlabeled data's label from two modal features by:
\begin{equation}
\begin{split}
        \left\{\begin{array}{ll}
            \mathbf{Y}_u^r = 
            {\mathbf{X}_u^r}^T \mathbf{P}^r  \\
            \mathbf{Y}_u^m = 
            {\mathbf{X}_u^m}^T \mathbf{P}^m  
        \end{array}\right.
\end{split}
\label{eqa: predict_unlabeled_data}
\end{equation}
where $\mathbf{Y}_u^r$ and $\mathbf{Y}_u^m$ denote predicted \textbf{soft-pseudo label matrices}  of unlabeled data on rotation-modal and mirror-modal.

(iii) Rank the values in soft-pseudo label matrices, then selecting the most confident unlabeled samples' feature $\mathbf{x}_{select}^r$ and $\mathbf{x}_{select}^m$ on each modal, then asserting them corresponding \textbf{one-hot-pseudo label vectors}  $\mathbf{y}_{select}^r$ and $\mathbf{y}_{select}^m$ 
(for more details about how to select the most confident sample, please refer to Section \ref{sec: select_most_confident_sample}).
Next, crossly extend the pseudo-labeled samples and corresponding labels to the support set on different modals. We formulate this step as:
\begin{equation}
\begin{split}
        \left\{\begin{array}{llll}
            \mathbf{X}_s^r = [\mathbf{X}_s^r, \mathbf{x}_{select}^m],
            \mathbf{Y}_s^r = [\mathbf{Y}_s^r, \mathbf{y}_{select}^m]\\
            \mathbf{X}_s^m = [\mathbf{X}_s^m, \mathbf{x}_{select}^r],
            \mathbf{Y}_s^m = [\mathbf{Y}_s^m, \mathbf{y}_{select}^r]\\
        \end{array}\right.
\end{split}
\label{eqa: extended_support_set}
\end{equation}
where $\mathbf{Y}_s^r$ and $\mathbf{Y}_s^m$ denote the \textbf{one-hot label matrices} of support data on two modals.

(iv) Repeat (i), (ii) (iii) until the unlabeled data is exhausted (in the real application, we usually select $80$ unlabeled samples, for more discussions and results, please see Section \ref{subsubsec: Influence of GCT Block} and Figure \ref{fig: ablation_studie_unlabeled_instance}).
Then we obtain two optimal classifiers $\mathbf{P}_{opt}^r$ and $\mathbf{P}_{opt}^m$. Employ them to predict the query labels by:
\begin{equation}
\begin{split}
    \mathcal{Z}(\mathbf{X}_q^r,\mathbf{X}_q^m) = id_{max} \left\{ 
    \frac{\left( {\mathbf{X}^r_{q}}^T \mathbf{P}_{opt}^r + {\mathbf{X}^m_{q}}^T \mathbf{P}_{opt}^m \right) }{2} \right\}
\end{split}
\label{eqa: predict_query_set_label}
\end{equation}
where 
$ \mathbf{Y}_q^r = {\mathbf{X}^r_{q}}^T \mathbf{P}_{opt}^r$ and $\mathbf{Y}_q^m = {\mathbf{X}^m_{q}}^T \mathbf{P}_{opt}^m$, denote the predicted \textbf{soft label matrices} of query data on rotation-modality and mirror-modality. 

Next see the transductive semi-supervised few-shot learning (TSSFSL). The query feature is also available when constructing the classifier. 
Here, we can implement TSSFSL with only a few minor tweaks.
Specifically, in step (ii), we have to predict not only unlabeled data but also query data.

Then see the transductive supervised few-shot learning (TSFSL), the unlabeled data is not available, but the query feature is given in advance. Therefore, we just need to replace the unlabeled data with query data in (i)(ii)(iii) steps, and finally classify the query data.

At last see the inductive supervised few-shot learning (ISFSL), the unlabeled data is unavailable, and query data is not given to us in advance.
Thus, the co-training strategy can not be used in this case and we think the basic IGL classifier is optimal. We can complete ISFSL by Equation \ref{eqa: two_model_base_learner} and \ref{eqa: predict_query_set_label}.

\subsection{How to Select the Most Confident Sample?}
\label{sec: select_most_confident_sample}

Here we take the rotation-modal as an example to illustrate the strategy. 
According to Equation \ref{eqa: predict_unlabeled_data}, we can get the predicted rotation-model soft label matrix $\mathbf{Y}_u^r \in \mathbb{R}^{N_u \times C}$, where $N_{u}$ denotes the number of unlabeled samples; $C$ denotes the number of unlabeled categories. 
For each element ${\mathbf{Y}_u^r}_{(n,c)}$, it means the probability of $n$-${th}$ sample belongs to the $c$-${th}$ category, where $n = 1,2,\cdots, N_u$, $c = 1,2,\cdots,C$. 
In our strategy, we first traverse all the elements in ${\mathbf{Y}_u^r}$ to find the largest element, which can be defined as ${\mathbf{Y}_u^r}_{(n_{max},c_{max})}$. 
The $n_{max}$-$th$ sample is the to-be-selected most confident sample, which belongs to the $c_{max}$-$th$ category.






\subsection{Multi-Modal Feature Extractor}
\label{subsec: FE}
We can adopt various modal features from different feature extractors to achieve our purpose.
For example:
(i)
Standard modality (Std-Mod) \cite{wang2020instance}, the feature extractor comes from a standard CNN-based classification structure.
(ii)
Meta modality (Meta-Mod) \cite{lee2019meta}, the feature extractor combines the strategy of meta-learning with the network.
(iii)
Self-supervised rotation modality (SS-R-Mod) \cite{mangla2020charting}, the feature extractor introduces auxiliary loss to predict the angle of image rotation, including $\{0^{\circ}, 90^{\circ}, 180^{\circ}, 270^{\circ} \}$.
(iv)
Self-supervised mirror modality (SS-M-Mod) \cite{shao2021mhfc}. Different from the SS-R-Mod, SS-M-Mod introduces another auxiliary loss to predict image mirrors, including $\{vertically, horizontally, diagonally \}$.
In most of the experiments, we present the results of collaborative training with SS-R-Mod and SS-M-Mod, and thereby we briefly introduce them. 

In the SS-R-Mod, the feature extractor updates the network parameters with two kinds of loss, which are the standard classification loss (\textit{i.e.}, $\mathcal{L}^{s}$) and rotation-based self-supervised auxiliary loss (\textit{i.e.}, $\mathcal{L}^{r}$). 
To be more specific, assume there's a base image feature vector $\mathbf{x}$. We project it into a label space, \textit{i.e.}, $\mathbf{x} \rightarrow \mathbf{z}^{s}$, where $\mathbf{z}^{s} = [z^{s}_1,z^{s}_2,\cdots,z^{s}_{C_{base}}] \in \mathbb{R}^{C_{base}}$, $C_{base}$ denotes the number of the base category. Then transform it to the probability distribution and calculate the standard classification loss $\mathcal{L}^{s}$ by cross-entropy function:
\begin{equation}
\begin{split}
    \mathcal{L}^{s}
    =-\sum_{c=1}^{C_{base}} \hat{y}^{s}_c log(y^{s}_c)
\end{split}
\label{eqa: standard_classification_loss}
\end{equation}
where 
$y^{s}_c = \frac{e^{z^{s}_c}}{\sum_{c=1}^{C_{base}} e^{z^{s}_c}}$ indicates the predicted probability that sample $\mathbf{x}$ belongs to class $c$, while $\hat{y}^{s}_c$ denotes the groundtruth probability.
After that, we map the $\mathbf{x}$ to rotation-based label space, \textit{i.e.}, $\mathbf{x} \rightarrow \mathbf{z}^{r}$, where $\mathbf{z}^{r} = [z^{r}_1,z^{r}_2,z^{r}_3,z^{r}_4] \in \mathbb{R}^{4}$. We can get the auxiliary loss by:
\begin{equation}
\begin{split}
    \mathcal{L}^{r}
    =-\sum_{c=1}^{4} \hat{y}^{r}_c log(y^{r}_c)
\end{split}
\label{eqa: rotation_self_supervised_loss}
\end{equation}
where $y^{r}_c = \frac{e^{z^{r}_c}}{\sum_{c=1}^{4} e^{z^{r}_c}}$
indicates the predicted probability of rotation angle, while $\hat{y}^{r}_c$ denotes the groundtruth probability. The complete loss in SS-R-Mod is $\mathcal{L}^{s} + \mathcal{L}^{r}$.

In the SS-M-Mod, it also uses the standard classification loss, but change the rotation-based self-supervised auxiliary loss to mirror-based self-supervised auxiliary loss $\mathcal{L}^{m}$, which can be defined as:
\begin{equation}
\begin{split}
    \mathcal{L}^{m}
    =-\sum_{c=1}^{3} \hat{y}^{m}_c log(y^{m}_c)
\end{split}
\label{eqa: mirror_self_supervised_loss}
\end{equation}
where $y^{m}_c$
indicates the predicted probability of mirroring way, while $\hat{y}^{m}_c$ denotes the groundtruth probability.
The complete loss in SS-M-Mod is $\mathcal{L}^{s} + \mathcal{L}^{m}$.

\subsection{Discussion and Analysis}
\label{sec: Discussion}
In this section, we will further discuss and analyze our work from two aspects. We first see the comparison of our Isolated Graph Learning (IGL) with traditional Graph Learning (GL); next conclude with a comprehensive analysis of the reasons for the success of GCT.

\subsubsection{Comparison of IGL and GL}
Compared with IGL, the GL's objective function is different, which can be formulated as:
\begin{equation}
\begin{split}
        \mathcal{F}(\mathbf{R}) = f_4(\mathbf{R}) + \beta f_5(\mathbf{R})
\end{split}
\label{eqa:GL_cost_function}
\end{equation}
where 
$\mathbf{R}$ is the to-be-predicted soft label matrix.
$\beta$ denotes the parameter to balance the function.
$f_4(\mathbf{R})$ denotes the graph Laplacian regularizer, which can be formulated as:
\begin{equation}
\begin{split}
        f_4(\mathbf{R}) 
        = \text{tr} \left( \mathbf{R}^T \mathbf{\Delta} \mathbf{R} \right)
\end{split}
\label{eqa:GL_graph_laplacian}
\end{equation}
where 
$\mathbf{\Delta}= \mathbf{D}^{- \frac{1}{2}} \mathbf{A} \mathbf{D}^{- \frac{1}{2}}$ denotes the normalized graph Laplacian operator.
$f_5(\mathbf{R})$ indicates the empirical loss term, which can be formulated as:
\begin{equation}
\begin{split}
        f_5(\mathbf{R}) = \left \| \mathbf{R} - \mathbf{Y} \right\|_F^2 
\end{split}
\label{eqa:GL_empirical_loss_term}
\end{equation}
where $\mathbf{Y}$ indicates the initial one-hot label matrix.

Comparing equations \ref{eqa:cost_function},\ref{eqa:graph_laplacian},\ref{eqa:empirical_loss_term} and equations \ref{eqa:GL_cost_function},\ref{eqa:GL_graph_laplacian},\ref{eqa:GL_empirical_loss_term}, it can be seen that the most essential difference between IGL and GL is that we replace $\mathbf{R}$ with $\mathbf{X}^T\mathbf{P}$. If directly using $\mathbf{R}$, the training samples and the to-be-classified testing samples must be employed together to achieve the testing samples' label prediction.
In other words, if we only use the training data to construct the graph, we can only achieve label propagation between training samples. When a new batch of testing data arrives, we must rebuild the graph based on the training and testing data to realize the label prediction for the testing samples.
While IGL is different. After we get $\mathbf{P}$ based on the training data, we can directly predict the label of new testing data based on $\mathbf{P}$ without rebuilding the graph.

\subsubsection{Comprehensive Analysis of GCT}
Here, let's sort out why GCT is effective. After dismantling it, we find that there are three parts that positively influence our method:

(i) The first point is that the base classifier IGL we designed maps the original features to the graph space, which can reduce the dependence on features in the FSL task, thereby weakening the influence of the FEM problem.

(ii) The second point is that we introduce multi-modal information. 
In the FSL task, the features obtained by different feature extractors are different. Although they all cause the distribution-shift, the angle of shift varies. Through the mutual correction of multi-modal information, the final performance can be improved.

(iii) The third point is that the employed co-training strategy is reasonable and efficient. As mentioned before, we use multi-modal information here. However, multi-modal features are a double-edged sword. If used well, features with different shortcomings can be corrected with each other to improve performance. If used incorrectly, the performance will be further degraded.
In the co-training strategy, we select the most confident sample in a single modality (the selected sample can be treated as the one that is not affected by distribution-shift), and then amplify the advantages of each modality through the alternate iterations of the two modalities, so as to enhance the classifier's ability.






\section{Conclusion}
There is a fundamental problem in Few-shot learning based tasks, \textit{i.e.}, Feature-Extractor-Maladaptive (FEM) problem. In this paper, we make two efforts to address this challenge. 
First, we propose a novel label prediction method, Isolated Graph Learning (IGL), to encode the feature embedding to graph representation and then propagate the label information through graph structure for prediction. Second, we extend IGL to the co-training framework to exploit multi-modal features in the semi-supervised setting, dubbed as Graph Co-Training (GCT). From the two perspectives, we have tackled this challenge to some extent. 
In our future work, we may study to improve the quality of the co-training strategy.

{
    \small
    \bibliographystyle{ieeenat_fullname}
    \bibliography{main}
}


\end{document}